\documentclass[final,3p,times,twocolumn]{elsarticle}
\usepackage{geometry}
\newdefinition{definition}{Definition}
\usepackage{amsmath}

\journal{}
\usepackage{bm}
\usepackage{mathrsfs}
\usepackage{amsfonts}
\usepackage{amssymb}
\usepackage{amsmath}
\usepackage{latexsym}
\usepackage{graphicx}
\usepackage{subfigure}
\usepackage{amsthm}
\usepackage{indentfirst}
\usepackage[linesnumbered,ruled]{algorithm2e}
\usepackage{multirow}
\usepackage{longtable}

\usepackage{graphics}
\usepackage{graphicx}
\usepackage{epstopdf}

\theoremstyle{definition}
\newtheorem{rem}{Remark}
\usepackage{graphicx}
\usepackage{amssymb}
\usepackage{amsthm}
\usepackage{amsmath}
\usepackage{lineno}
\usepackage{txfonts} 
\usepackage{enumitem}
\usepackage{multirow}
\usepackage{caption}
\usepackage{txfonts}
\usepackage{array}
\usepackage{booktabs}
\usepackage{longtable}
\usepackage{bm}
\usepackage{setspace}
\usepackage{natbib}
\usepackage{xcolor}
\usepackage{color} 
\usepackage{adjustbox}

\begin{document}

\begin{frontmatter}

\title{Optimal Gait Design for a Soft Quadruped Robot via Multi-fidelity Bayesian Optimization}

\author[1address]{Kaige Tan}
\author[1address]{Xuezhi Niu}
\author[1address]{Qinglei Ji}
\author[1address]{Lei Feng\corref{mycorrespondingauthor}}
\cortext[mycorrespondingauthor]{Corresponding author}
\ead{lfeng@kth.se}

\author[1address]{Martin T{ö}rngren}

\address[1address]{Department of Engineering Design, KTH Royal Institute of Technology, Stockholm 10044, Sweden}

\begin{abstract}
This study focuses on the locomotion capability improvement in a tendon-driven soft quadruped robot through an online adaptive learning approach. Leveraging the inverse kinematics model of the soft quadruped robot, we employ a central pattern generator to design a parametric gait pattern, and use Bayesian optimization (BO) to find the optimal parameters. Further, to address the challenges of modeling discrepancies, we implement a multi-fidelity BO approach, combining data from both simulation and physical experiments throughout training and optimization.  This strategy enables the adaptive refinement of the gait pattern and ensures a smooth transition from simulation to real-world deployment for the controller.
Moreover, we integrate a computational task off-loading architecture by edge computing, which reduces the onboard computational and memory overhead, to improve real-time control performance and facilitate an effective online learning process. The proposed approach successfully achieves optimal walking gait design for physical deployment with high efficiency, effectively addressing challenges related to the reality gap in soft robotics.
\end{abstract}


\end{frontmatter}


\section{Introduction}
\label{sec:introduction}
Soft robotics, utilizing compliant materials and structures as actuators, is gaining prominence for creating adaptive and resilient robotic systems, particularly in soft quadruped robots with legged locomotion~\cite{yasa2023overview}. In contrast to their rigid counterparts, soft quadruped robots are featured with compliant materials and soft actuators, emulating the flexibility of biological tissue. Actuated by pneumatic pressure~\cite{diteesawat2021electro, tan2022shape}, electric voltage~\cite{ma2022review, ji2022synthesizing}, and magnetic fields~\cite{ebrahimi2021magnetic, dong2022untethered}, soft robots demonstrate resilience to diverse working conditions, enhance safety in human-machine interactions, and effectively absorb external disturbances through their inherent elasticity~\cite{ji2022omnidirectional}.

While soft actuators facilitate intricate deformations for complex locomotion, effective gait design is the necessity of exploiting the full walking potential of soft robots~\cite{drotman2021electronics}. 
The challenges arise from the soft robots' modeling complexity under environmental interactions and the inherent difficulty of maintaining consistent actuation force and precision for soft actuators~\cite{tan2023edge}, a concern not shared by their rigid counterparts.

To resolve the above challenge, tendon-driven soft actuators (TSAs) are utilized for actuation, where the actuators' bending and compression motions are synchronized to improve the walking performance of the robot~\cite{ma2019first, drotman2021electronics, ji2022omnidirectional, wensing2023optimization}. 
Prior research primarily focuses on predefined gait trajectories~\cite{drotman2021electronics, ji2022omnidirectional} or model-based control designs for gait optimization~\cite{ma2019first, wensing2023optimization}. However, the gait trajectory is either not optimized or only optimized through Finite Element Analysis (FEA)~\cite{bern2019trajectory, 10136424} which fails to achieve good time efficiency. Therefore, a research gap persists for advanced gait exploration in developing an optimal gait pattern for TSA-type soft robots.

To create adaptive and efficient locomotion patterns, many state-of-the-art techniques provide sophisticated solutions tailored for rigid quadruped robots, including 
reinforcement learning~\cite{ibarz2021train, miki2022learning, 9693519}, Central Pattern Generators (CPGs)~\cite{LAN2021107688, 9932888, zhang2023hybrid}, and model predictive control~\cite{ding2021representation, gao2022trajectory, 10138309}. A comprehensive overview of the relevant approaches can be found in the work of Taheri et al.~\cite{taheri2023study}. 
Specifically, artificial CPGs, which contain limited cycles with repeating patterns for gait synchronization, have been proposed as a robust approach to enhancing gait stability. CPGs can be designed by modeling the behavior of neuron cells as neuron oscillators~\cite{hurkey2023gap}. Alternatively, another type of CPG is designed based on dynamic systems, e.g., Hopf oscillators~\cite{shao2021learning} and van-der-Pol oscillators~\cite{song2023gaits}. The oscillators can generate rhythmic signals in the form of trajectories and represent the stance-swing sequence of the legs with continuous states, making them well-suited for modeling gait patterns in soft quadruped robots.

Essentially, these oscillator models often contain tuning parameters, which require a well-designed method for users to properly choose their values. There are two major considerations for achieving optimal performance when applying the parametric locomotion design framework to the soft quadruped robots:
\begin{itemize}
    \item \textbf{Uncertain parameter optimization}: Owing to the intrinsic nonlinear and time-variant properties of the soft material, the dynamics of the soft actuators is subject to significant uncertainty. Traditional deterministic functions and optimization methods lack robustness when applied to the soft robot gait design.
    \item \textbf{Modeling error compensation}:
    Precisely modeling the dynamics of soft materials is challenging due to their compliant nature. Consequently, it is imperative to address the discrepancies between simulation and real-world performance of soft robots. The inherent complexities of soft robotic systems pose challenges for simulators to accurately replicate them. Therefore, it becomes crucial to devise strategies for reducing this modeling error during physical deployment.
\end{itemize}

To address the first challenge, the implicit relationship between the model parameters and the objective function is generally treated as a stochastic black-box model, where optimal parameter values can be determined through black-box optimization methodologies. Solving black-box optimization via soft computing has proven effective in various robotics applications. Notably, algorithms such as Bayesian Optimization (BO)~\cite{lan2021learning}, Particle Swarm Optimization (PSO)~\cite{okulewicz2022self}, and Evolutionary Algorithms (EA)~\cite{nadizar2023experimental} have been extensively employed. Among them, BO has the advantage of considering probabilistic objective functions and using efficient sampling to achieve fast convergence. Thus, given the parametric-based gait pattern, BO stands out as an ideal choice for systematically tuning uncertain parameters, thereby unlocking the potential of the robot's gait with high exploration efficiency.

The second challenge stems from the inevitable discrepancies between the simulation model and the real robot, owing to noise, perturbations, and modeling errors. This disparity, often referred to as the \emph{reality gap}, leads to discrepancies within a simulator-aided framework, necessitating adjustments to the optimal solution identified in the simulation environment through physical experiments~\cite{jiang2023stable}.
When developing a data-driven controller, abundant synthetic training data can be obtained from the simulator, but the data contain unavoidable errors compared to data derived from real robots. Moreover, while measurement data from real robots are of high quality, obtaining them in large quantities is challenging due to the need for human supervision and experiment resets, resulting in a slow and costly process~\cite{ibarz2021train}. To address this challenge, our method mixes the two types of data in a single learning algorithm to alleviate the reality gap between simulation and physical robots.
Specifically, we focus on the BO method and propose a \emph{multi-fidelity BO} (MFBO) approach to achieve efficient sim-to-real adaptation of the optimal points between the model and the robot. Consequently, the optimal walking pattern learned from simulations can be efficiently adapted to the physical robot through online learning in the real world.

MFBO extends the traditional BO framework to incorporate information from diverse sources with different fidelity levels. While BO typically relies on a Gaussian Process (GP) model as a surrogate to represent objective function evaluations, it is limited by a scalar output for a single model characterization.
To address this limitation, either separate GPs are used for distinct models~\cite{perdikaris2016model}, or an error function is trained to delineate the difference among different models~\cite{marco2017virtual}. Additionally, the multi-task GP (MTGP) method~\cite{wei2021review, zhang2021survey} has been introduced to naturally manage multiple coupled outputs, capturing complex data dependencies arising from various sources. 

Incorporating MTGP into the BO framework gives rise to the MFBO method, which facilitates accelerated model training across different data sources. Recently, MFBO has found applications in various research domains, e.g., hyper-parameter optimization~\cite{horvath2021hyperparameter}, process optimization~\cite{wang2020transfer}, and safe controller design~\cite{lau2023multi}. However, its potential for modeling error compensation in robotics systems, particularly in adaptive controller design, remains underexplored. 

Similar to this paper, Lan et al.~\cite{LAN2021107688} integrate BO and CPG for the robot's optimal gait pattern generation. Although they identify the reality gap during the physical testing, they do not propose a method to adapt the gait pattern to the physical robot.
The shortcoming prompts our exploration of the MFBO approach 
to the efficient adaptation of the gait controller learned from simulations to the physical robot. The adaptation procedure is performed through online learning with the real robot, which combines both previously obtained simulation data and the online generated real-world data to obtain a better gait controller for the physical robot.
It is worth noting prior studies do not distinguish the terms \emph{multi-fidelity} and \emph{multi-task} within the same BO framework. We adopt the term \emph{multi-fidelity} in the sequel to maintain clarity and underscore its relevance within the context of sim-to-real transfer scenarios~\cite{fernandez2023review}.

Moreover, the complexities of online MFBO training and the challenges associated with CPG gait generation pose difficulties for achieving real-time control on the robot’s onboard computation system. Recently, the integration of edge computing has emerged as a promising solution to enhance robotic system performance~\cite{sanchez2022edge, groshev2023edge}. 
Computational task off-loading becomes instrumental in load balancing~\cite{yang2022inverse} and optimizing resource utilization~\cite{tan2022decentralized}. Furthermore, compared to task off-loading to the public cloud, the integration of edge computing allows for data processing closer to the source with less latency, ensures real-time computational task completion~\cite{tan2023edge}, and improves the system responsiveness~\cite{qiu2020edge}. 
Therefore, our study leverages the computational off-loading enabled by edge computing to improve the real-time control performance for the soft quadruped robot gait design. 

In summary, the contributions of our article are as follows.
\begin{enumerate}
    \item A gait generator is devised for the soft quadruped robot with CPG-based modeling. The utilization of a parametric formulation enhances the adaptability and flexibility of gait pattern design.
    \item The computation of the optimal gait pattern parameters is formulated as a stochastic black-box optimization problem, which is solved using the MFBO approach. Our approach effectively mitigates the reality gap and achieves a rapid transition from sim-to-real adaptation.
    \item An online learning architecture is designed based on an edge computing framework. Complex MFBO optimization and training tasks are performed at the edge server. Computational task off-loading, facilitated by 5G communication, allows for real-time adaptive parameter optimization and fast deployment.
\end{enumerate}

The rest of this article is organized as follows. Followed by the preliminaries in Sec.~\ref{sec:preliminaries}, Sec.~\ref{sec:Gait_Pattern} introduces the soft quadruped robot under study and CPG-based gait pattern model design. Then, the formulated parameter optimization problem is solved by MFBO in Sec.~\ref{sec:online_adaption}. Sec.~\ref{sec:experiment_setup} describes the hardware implementation and the experiment setup. Sec.~\ref{sec:result_validation} shows the evaluation and results of the approach. Finally, Sec.~\ref{sec:conclusion} concludes this article.

\section{Preliminaries}
\label{sec:preliminaries}
\subsection{Gaussian Processes Model}
\label{sec:GP}
A GP model represents a stochastic function $g\!:\!\mathcal{P}\!\rightarrow\!\mathbb{R}$, where $\mathcal{P}$ is a subset of $\mathbb{R}^n$. A GP model is specified by its mean and covariance functions, i.e., $m\!:\!\mathcal{P}\!\rightarrow\!\mathbb{R}$ and $k\!:\!\mathcal{P}\times\mathcal{P}\!\rightarrow\!\mathbb{R}$. For a point $\mathbf{p} \in \mathcal{P}$, $g(\mathbf{p}) \sim \mathcal{GP}\big(m(\mathbf{p}), k(\mathbf{p}, \mathbf{p})\big)$ is a random value with the Gaussian distribution whose mean is $m(\mathbf{p}) \!=\! \mathbb{E}\left[g(\mathbf{p})\right] \!\in\! \mathbb{R}$ and variance is $k(\mathbf{p}, \mathbf{p})$. For any two points $\mathbf{p}, \mathbf{p}' \in \mathcal{P}$, the covariance is defined by $k(\mathbf{p}, \mathbf{p}') \!=\! \mathbb{E}\left[\big(g(\mathbf{p}) \!-\! m(\mathbf{p})\big) \big(g(\mathbf{p}') \!-\! m(\mathbf{p}')\big)\right] \!\in\! \mathbb{R}_{+}$.
The covariance function defines the relationships between data points in the input space, which is characterized by the hyperparameters $\pmb{\lambda}$. They impact the shape of regression functions and can be optimally defined by log marginal likelihood. With a training data set $\mathcal{T} = \{[\mathbf{p}_i, v_i], 
 i\!=\!1,\dots,n\}$ of $n$ training samples $\mathbf{P} = [\mathbf{p}_1,\dots, \mathbf{p}_{n}]$ and their corresponding outputs $\mathbf{v} = [v_1, \dots, v_n]$, the optimal values $\pmb{\lambda}^*$ can be acquired by maximizing the likelihood of the observed outputs. Based on that, the GP model can be used to predict the function value $v'$ of a test point $\mathbf{p}'$, yielding
\begin{equation}
\label{eq:posterior_likelihood}
    p(v' \mid \mathbf{p}', \pmb{\lambda}^*, \mathcal{T}) = \mathcal{N}\left(\mu_{\mathbf{p}'}, \sigma_{\mathbf{p}'}\right),
\end{equation}
and
\begin{equation}
\label{eq:mean_covariance_expression_actuator}
\begin{aligned}
    & \mu_{\mathbf{p}'} = k(\mathbf{P}, \mathbf{p}')^\top K_\mathbf{P}^{-1} \mathbf{v}^\top,\\
    & \sigma_{\mathbf{p}'} = k(\mathbf{p}', \mathbf{p}') - k(\mathbf{P}, \mathbf{p}')^\top K_\mathbf{P}^{-1} k(\mathbf{P}, \mathbf{p}'),
\end{aligned}
\end{equation}
where $\mu_{\mathbf{p}'}$ and $\sigma_{\mathbf{p}'}$ are the mean and variance of $v'$. $K_\mathbf{P}$ is the symmetric and positive semi-definite $n \times n$ covariance matrix and $K_\mathbf{P}(i, j) = k(\mathbf{p}_{i}, \mathbf{p}_{j})$. $k(\mathbf{P}, \mathbf{p}')$ is an $n \times 1$ vector, and each entry denotes the kernel function value between $\mathbf{p}_i \in \mathbf{P}$ and $\mathbf{p}'$. The values of $k(\cdot, \cdot)$ and $K_\mathbf{P}$ in (\ref{eq:mean_covariance_expression_actuator}) depend on $\pmb{\lambda}^*$.

\subsection{Bayesian Optimization}
\label{sec:Bayesian_opt}
Bayesian optimization (BO) is an effective optimization technique for black-box optimization. The method employs a data fitting surrogate model $g(\cdot)$ to approximate the unknown objective function $J(\cdot)$. The selection of the surrogate model is crucial for accuracy and efficiency~\cite{bliek2023benchmarking}. Popular surrogate models include GP, random forest (RF), piece-wise linear (PL), and etc.. The GP model is often preferred for two primary reasons: 1) its ability to quantify uncertainty in predictions, and 2) its adaptation for mitigation of the reality gap in MFBO. Comparisons of different surrogate models for BO are presented later in Sec.~\ref{sec:surrogate_comp}.

Based on the training data set and the GP model hyperparameters, Bayesian optimization maximizes an acquisition function $a(\cdot)$ to determine the next most promising point to evaluate in the search for global optimality of $J(\cdot)$, i.e., $\mathbf{p}_\textrm{next}\!=\!\textrm{argmax}_\mathbf{p} \, a(\mathbf{p}\mid\mathcal{T}, \pmb{\lambda})$.
After evaluating the newly generated point, i.e., obtaining $J(\mathbf{p}_\textrm{next})$, the GP model will also be updated based on the new observation. The iterative cycle continues until predetermined stopping conditions are met, such as a predetermined number of iterations or convergence criteria. If the stopping conditions are proper, the iteration converges to the optimal solution.


Many types of acquisition functions $a(\cdot)$ have been proposed, e.g., expected improvement criterion (EI), Probability of Improvement (PI), and GP upper
confidence bounds (GP-UCB). EI has proved its superiority over PI, and it does not require the tuning parameter as in GP-UCB~\cite{snoek2012practical}. Thus, EI is adopted in this study, which maximizes the expected improvement over the current best $J^*$, i.e., 
\begin{equation}
    \label{eq:EI_acquisiton}
    a_\textrm{EI}(\textbf{p}\mid \mathcal{T}, \pmb{\lambda}) = \mathbb{E}(\max[0, g(\mathbf{p})\!-\!J^*]\mid \mathcal{T}, \pmb{\lambda}).
\end{equation}
It has a closed form under the GP surrogate model, i.e.,
\begin{equation}
\label{eq:EI_acquisition_fun}
\begin{aligned}
    &a_\textrm{EI}(\mathbf{p}\mid\mathcal{T}, \pmb{\lambda}) = \sqrt{\sigma_\mathbf{p}}\big[\gamma(\mathbf{p}) \Phi\big(\gamma(\mathbf{p})\big) + \phi(\mathbf{p})\big],\\
    &\gamma(\mathbf{p}) = (\mu_\mathbf{p}-J^*)/\sqrt{\sigma_\mathbf{p}},
\end{aligned}
\end{equation}
where $\mu_\mathbf{p}$ and $\sigma_\mathbf{p}$ are obtained from (\ref{eq:mean_covariance_expression_actuator}). $\phi(\cdot)$ denotes the standard normal density function, and  $\Phi(\cdot)$ denotes the cumulative distribution function of the standard normal distribution.

\subsection{Hopf Oscillator}
\label{sec:hopf_oscillator}
As a method to model the artificial CPG, Hopf oscillator produces various harmonic output patterns, and has a clear correlation with the applied coefficients. It can be denoted by the oscillator state $\mathbf{o}$, which takes the form of the nonlinear differential equations:
\begin{equation}
\label{eq:hopf_oscillator}
\Dot{\mathbf{o}} = f_\textrm{h}(\mathbf{o}) + \mathbf{q}
 = \begin{bmatrix}
        k(A^2-o_1^2-o_2^2)o_1 - 2\pi fo_2\\
        k(A^2-o_1^2-o_2^2)o_2 + 2\pi fo_1
    \end{bmatrix} + \begin{bmatrix}
        q_{1}\\
        q_{2}
    \end{bmatrix},
\end{equation}
where $\mathbf{o} = \left[o_1, o_2\right]^\top$ is the Hopf oscillator state vector. $\mathbf{q}\!=\!\left[q_{1}, q_{2}\right]^\top$ is a design coupling vector to coordinate different oscillators. The formulas for $q_1$ and $q_2$ are defined later in (\ref{eq:coupling_term}). $A, f$, and $k$ are design parameters that determine the amplitude of the steady-state oscillation, oscillation frequency, and the speed of convergence, respectively. Given the design parameters and a non-zero initial state, a harmonic pattern will be generated from the oscillator represented by $o_1$ and $o_2$. Note that $\mathbf{o}$ is time-dependent; however, for the sake of simplicity, we shall include the argument of time in the expression only when essential.

In addition, a dynamic modulation of frequency is introduced to generalize the oscillator formulation. 
It is realized by employing a varying frequency influenced by the state of oscillator, thus achieving a non-harmonic pattern.
Given the shape ratio $\alpha$ and time constant $\tau$, the constant frequency $f$ in (\ref{eq:hopf_oscillator}) can be adjusted to yield a frequency $f_a$, where
\begin{equation}
    \label{eq:non_harmonic_freq}
    f_a(o_2) = \frac{f}{2\alpha} + \frac{(2\alpha-1)f}{2\alpha(1-\alpha)(1+e^{-\tau o_2})}.
\end{equation}
Note that the period $T$ of the oscillator in (\ref{eq:non_harmonic_freq}) is still determined by $f$ (i.e., $T = 1/f$), and the harmonic pattern is a special case of (\ref{eq:non_harmonic_freq}) when $\alpha=0.5$. When $\alpha \neq 0.5$, the frequency $f_a$ changes monotonically in each oscillation period. We direct the interested reader to reference~\cite{zhou2011design} for the comprehensive analysis on Hopf oscillators. Note that $f_a$ is a state-dependent variable of $o_2$. As a result, the other state, i.e., $o_1$, becomes the exclusive determinant of the fundamental gait trajectory, and its value serves as the output of the Hopf oscillator in the sequel.

\section{Parametric Gait Pattern Model}
\label{sec:Gait_Pattern}
\subsection{Inverse Kinematics: Task Level to Motor Level}
\label{sec:inverse_kinematics}
\begin{figure}[htb]
	\centering
		\includegraphics[width=3in]{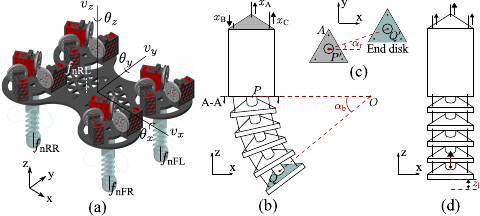}
	\caption{Overview of the soft robot and its Tendon-driven Soft Actuator (TSA): (a) Rendered robot with key states. (b) Structure of TSA: rigid upper thigh and bendable lower section with $\alpha_b$. (c) Top view of bent TSA with $\alpha_r$. (d) TSA compression via equal tendon pull with $z_\textrm{l}$.} 
    \label{fig:robotsmodel}
\end{figure}

Our previous works~\cite{ji2022synthesizing, ji2022omnidirectional} introduce the design and fabraction of a soft quadruped robot characterized by its innovative utilization of four TSAs. Figure~\ref{fig:robotsmodel}(a) provides a visual representation of the robot, illustrating key states of robot such as roll ($\theta_x$), pitch ($\theta_y$), yaw ($\theta_z$), translational velocities along three axes ($v_x,v_y,v_z$), and the normal contact forces on each foot ($f_{nFL},f_{nFR},f_{nRL},f_{nRR}$). The defining characteristic of the TSAs is the cable-driven mechanism, which enables seamless and continuous bending and twisting along its entire length. The cables, also referred to as tendons, guide the deformation of soft materials, facilitating precise movement to designated positions. Each leg of the robot is actuated by three cables, which are driven by servo motors, resulting in the generation of tendon displacement denoted as $\mathbf{x}_\textrm{ten} = \left[x_\textrm{A}, x_\textrm{B}, x_\textrm{C}\right]^\top$, as illustrated in Figure~\ref{fig:robotsmodel}(b). 
The state of each soft leg is characterized by 
\begin{equation}
\label{eq:robot_state}
\mathbf{s} = \left[ \alpha_\textrm{b}, \alpha_\textrm{r}, z_\textrm{l} \right]^\top,
\end{equation}
as shown in Figure~\ref{fig:robotsmodel}(b)-(d). The mapping of a feasible $\mathbf{s}$ into the corresponding tendon displacement $\mathbf{x}_\textrm{ten}$ is investigated in the previous work~\cite{ji2022omnidirectional} as the following function.
\begin{equation}
    \label{eq:inverse_kinematic}
    \mathbf{x}_\textrm{ten} = g_\textrm{inv}(\mathbf{s}) = f_\textrm{inv}([\alpha_\textrm{b}, \alpha_\textrm{r}]^\top) + \textbf{1} z_\textrm{l},
\end{equation}
where $f_\textrm{inv}: \mathbb{R}^2 \rightarrow \mathbb{R}^3$ is derived through geometric analysis, and
\begin{equation}
\label{eq:geometric_analysis}
\begin{aligned}
    &x_\textrm{A} = R_\textrm{d} \alpha_\textrm{b} \cos(\alpha_\textrm{r}),\\
&x_\textrm{B} = R_\textrm{d} \alpha_\textrm{b} \cos(\alpha_\textrm{r}+\frac{2\pi}{3}),\\
&x_\textrm{C} = R_\textrm{d} \alpha_\textrm{b} \cos(\alpha_\textrm{r}+\frac{4\pi}{3}).
\end{aligned}
\end{equation}
$R_\textrm{d}$ is the radius of the circle formed by the three guiding holes on a disk. $\mathbf{x}_\textrm{ten}$, as the tendon displacement, is further used as the motor reference for the low-level position control. The synchronization of the designed trajectory on $\mathbf{s}$ and $\mathbf{x}_\textrm{ten}$ will lead to a gait pattern of the robot on the task level.

Moreover, a simulation environment utilizing MATLAB Simulink Multibody blocks is developed~\cite{ji2022synthesizing, ji2022omnidirectional}, providing a digital emulation of the real robot. The plant model within the simulator exhibits a high level of modeling accuracy corresponding to the physical world, which supports and prompts the design of motion controllers. In addition, note that various terms, e.g., \emph{real robot}, \emph{physical robot}, and \emph{real-world robot}, are interchangeable in what follows, all referring to the entity existing physically within the tangible environment, as opposed to the simulated robotic platform employed.

\subsection{Gait Pattern Generation}
\label{sec:gait_traj_gen}
In the previous study~\cite{ji2022omnidirectional}, the temporal trajectory of $\mathbf{s}$ is characterized by an ad hoc design approach, which is non-optimal in terms of robot walking speed and efficiency. State-of-the-art research usually identifies the walking of a quadruped robot by rhythmic patterns; hence, the nonlinear oscillator can be used as an ideal option to model a rhythmic movement in a formalized and parametric way. Therefore, to improve the gait design, we employ nonlinear oscillators to model the rhythmic movement of the actuator states.

Using the inverse kinematic model in (\ref{eq:inverse_kinematic}), this study applies the Hopf oscillator as a fundamental pattern generator for the robot locomotion pattern design. Following the notation in Sec.~\ref{sec:hopf_oscillator}, we denote by $\mathbf{o}_i$ the oscillator state of cycle behavior $\mathbf{s}_i$ of leg-$i$, $i \in \mathcal{I} = \{1,2,3,4\}$. To unify the cycle behavior of each leg and promote their coordination, we assume the oscillator motion of legs $\mathbf{o}_i$ is originated from a primitive oscillator $\mathbf{o}$ based on (\ref{eq:hopf_oscillator})-(\ref{eq:non_harmonic_freq}). 

Nevertheless, the Hopf oscillators of each leg are yet coordinated to shape gait patterns and generate motor trajectory references. To realize it, coupling term $\mathbf{q}_i$ is designed to generate coupled oscillator $\mathbf{o}_i$ of each leg from the primitive oscillator $\mathbf{o}$.
The distinct behavior of each leg is represented by a phase difference 
\begin{equation}
\label{eq:theta_def}
    \pmb{\theta} = [\theta_1, \theta_2, \theta_3, \theta_4].
\end{equation}
Thus, the rhythmic trajectory of each leg is defined by
\begin{equation}
    \label{eq:coupling_oscillator}
    \mathbf{o}_i(t) = \mathbf{o}\left(t - \Delta t_i\right), \,\, \Delta t_i=\theta_i/(2\pi), \,\, i \in \mathcal{I}.
\end{equation}

\begin{figure}[!t]
	\centering
		\includegraphics[width=3in]{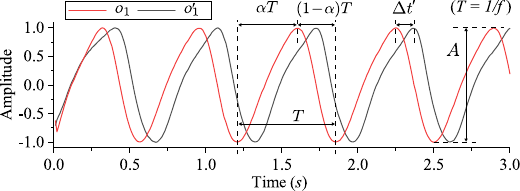}
	\caption{The non-harmonic oscillator response with varying frequencies. In this example, the initial state is randomized, and $[A,f,\alpha,\tau,k,\epsilon,\Delta t']\!=\![1,1.33,0.75,1,1000,5,0.25].$} \label{fig:oscillator_illustrate}
\end{figure}

This study considers that the oscillator $\mathbf{o}_i$ is coupled and perturbed by the primitive oscillator $\mathbf{o}$. Given the target phase difference $\theta_i = \pmb{\theta}[i]$, the coupling term of the $i$-th oscillator is calculated by
\begin{equation}
    \label{eq:coupling_term}
    \mathbf{q}_i = \begin{bmatrix}
        q_{1}\\
        q_{2}
    \end{bmatrix}_i = \begin{bmatrix}
        0\\
        \varepsilon(o_1\sin\theta_i + o_2\cos\theta_i)
    \end{bmatrix},
\end{equation}
where $\varepsilon > 0$ denotes the coupling strength. By designing the values of $\pmb{\theta}$, the oscillator of each leg $\mathbf{o}_i$ is defined. Thus, actuators of legs can be synchronized to compose different locomotive patterns, e.g., trotting, galloping, depending on the values of $\pmb{\theta}$. Figure~\ref{fig:oscillator_illustrate} illustrates an example of the non-harmonic oscillator response with varying frequencies, where $\mathbf{o}'$ is generated from $\mathbf{o}$. $o_1 = \mathbf{o}[1]$ and $o'_{1} = \mathbf{o}'[1]$ defined in (\ref{eq:hopf_oscillator}) are used as the output of the oscillator. In the figure, two coupled oscillators have the same pattern but different phases. 
$\Delta t'$ is defined by $\theta' = \pi/2$.

Subsequently, the derivation of $\mathbf{o}_i$ is applied to the design of $\mathbf{s}_i$. Recap that in $\mathbf{s}_i$, the rotational angle $\alpha_{r,i}$ of the leg influences the movement direction of the quadruped robot. Especially, the prior study~\cite{ji2022omnidirectional} has demonstrated that the direction of movement $\theta_{z}$ corresponds to the angular rotations of the legs when all four legs exhibit uniform values, i.e., $\forall i\in \mathcal{I}, \alpha_{r,i} = \theta_{z}$. Therefore, given a reference trajectory with the direction of movement $\theta_\textrm{ref}$, we can simplify the problem by setting $\alpha_{r,i} = \theta_\textrm{ref}$.

Thus, the cyclic behavior $\mathbf{s}_i$ is essentially defined by $\alpha_{b,i}$ and $z_{\textrm{l},i}$, which determine the trajectory and amplitude of the foot pattern. Based on the previous analysis, the rhythmic behaviors can be effectively characterized by the state output $o_{1,i}$ originating from the Hopf oscillator.  Note that, for the sake of simplifying the design process and limiting the parameter space, we consider $\alpha_{b,i}$ and $z_{\textrm{l},i}$ as interrelated, and both are characterized by the same oscillator output $o_{1,i}$ but with distinct phases and gains, i.e.,
\begin{equation}
    \label{eq:correlated_b_z}
    \alpha_{b,i}(t) = o_{1,i}(t), \quad \frac{\alpha_{b,i}(t)}{A_{\alpha_\textrm{b}}} = \frac{z_{\textrm{l},i}(t - \varphi)}{A_{z_\textrm{l}}}, 
\end{equation}
where $\varphi \in [0, T]$ is a design parameter, which represents the delay between the patterns of $\alpha_{b,i}$ and $z_{\textrm{l},i}$. This delay parameter allows for fine-tuning the synchronization between these two critical components in the gait pattern generation. $A_{\alpha_\textrm{b}}$ and $A_{z_\textrm{l}}$ denote the amplitudes for $\alpha_\textrm{b}$ and $z_\textrm{l}$, respectively. We consider that both trajectories share the same frequency $f$ and shape ratio $\alpha$ in order to generate the rhythmic pattern.

\subsection{Optimization Problem Formulation}
Based on the above analysis, the coordinated cycle behavior of each leg $\mathbf{s}_i$ can be defined by the oscillator state $\mathbf{o}_i$, the correlated phase difference $\varphi$, and the reference direction $\theta_\textrm{ref}$, and $\mathbf{o}_i$ is further influenced by the base oscillator $\mathbf{o}$ and coupling term $\mathbf{q}_i$ through (\ref{eq:coupling_oscillator})-(\ref{eq:correlated_b_z}). Thus,
\begin{equation}
    \label{eq:oscillator_to_cycle}
    \mathbf{s}_i = f_\textrm{CPG}\left(\mathbf{o}, \mathbf{q}_i, \varphi, \theta_\textrm{ref}\right).
\end{equation}
Eq.~(\ref{eq:oscillator_to_cycle}) provides a formulation of the rhythmic and coordinated gait pattern enabled by Hopf oscillator. This formulation enables the generation of coordinated leg movements in response to the specified reference direction and desired phase synchronization. Accordingly, the references $\mathbf{x}_\textrm{ten}$ for the 12 motors can be calculated from $\mathbf{s}_i$ through (\ref{eq:inverse_kinematic}). 
If the reference trajectories $\mathbf{x}_\textrm{ten}$ are provided to the robot's simulation model, the simulation outcomes can be analyzed to evaluate the quality of the gait pattern.

Generally, a gait of the leg robot is determined by three factors: stride, duty cycle, and phase factor. In our model, the phase factor is controlled by $\pmb{\theta}$ in (\ref{eq:theta_def}) for the leg synchronization, and the other two are derived from the trajectory of $\mathbf{s}_i$ based on (\ref{eq:oscillator_to_cycle}). Note that in (\ref{eq:oscillator_to_cycle}), given the initial state $\mathbf{o}(t_0)$, $\mathbf{o}$ is determined by parameters $A_{\alpha_\textrm{b}}, A_{z_\textrm{l}}, f, \alpha, \tau, k$, and $\mathbf{q}_i$ is computed based on $\varepsilon, \theta_i$. Among those, $\tau, k, \varepsilon$ control the stability and convergence of the oscillators but will not influence their steady-state behaviors; thus, those parameters are irrelevant to the gait design, which are regarded as constant and predefined in the design phase.

Therefore, besides $\theta_\textrm{ref}$ and $\pmb{\theta}$ that explicitly define the robot walking direction and phase factor (which can be pre-designed), the other walking properties are determined by $\mathbf{o}$, $\mathbf{q}_i$ and $\varphi$. To summarize, based on (\ref{eq:hopf_oscillator}), (\ref{eq:non_harmonic_freq}), and (\ref{eq:correlated_b_z}), the choice of parameters in the vector $\mathbf{p} = \left[A_{\alpha_\textrm{b}}, A_{z_\textrm{l}}, f, \alpha, \varphi\right]^\top \in \mathbb{R}^{5}$ significantly influences the stride and duty cycle of the gait, thus influences various aspects of the robot's walking properties, e.g., walking speed, Cost of Transport (COT), terrain clearance, and stride length.

By designing and optimizing the associated design parameters in $\mathbf{p}$, the parametric formulation of the gait pattern $\mathbf{s}_i$ enables the shaping of an optimal robot walking pattern. Considering the inverse kinematic mapping in Sec.~\ref{sec:inverse_kinematics}, the relationship between $\mathbf{p}$ and the objective function set up an end-to-end design from motor actuation commands to the robot walking task-level control. Specifically, as an intuitive representative of the interested properties, this study defines the objective function by maximizing the robot walking speed $v$. Nevertheless, the proposed approach can also be applied and analyzed for the other walking properties as mentioned before. Within our CPG-based parametric gait pattern framework, we assume that a set of parameter $\mathbf{p}$ 
will lead to a stabilized robot walking speed $v$. 

Owing to the lack of a known mathematical form, the evaluation of the walking speed $v$ for a candidate parameter vector $\textbf{p}$ is based on the simulation results of the robot model. If we view the simulated robot as a black-box model, this parameter design problem becomes a black-box optimization problem.
Given the unknown mapping function $J: \mathbb{R}^{5} \rightarrow \mathbb{R}$ with the design space $\mathcal{P}\!=\!\{\mathbf{p} \in \mathbb{R}^{5}: \underline{p_i} \leq \mathbf{p}_i \leq \overline{p_i}\}$, where $\underline{p_i}$ and $\overline{p_i}$ denote the lower and upper bounds for each element, the optimal walking speed can be identified through
\begin{equation}
    \label{eq:BO_objective_fun}
    v^* = \max_{\mathbf{p}\in\mathcal{P}} \,\,\, J(\mathbf{p}),
\end{equation}
This black-box optimization problem is solved by the Baysian optimization method introduced in Section~\ref{sec:Bayesian_opt}.

\section{Adaptive Optimal Gait Design by Multi-fidelity Bayesian Optimization}
\label{sec:online_adaption}

\subsection{Reality Gap from the Digital Replica}
\label{sec:reality_gap}
The optimization problem in (\ref{eq:BO_objective_fun}) can be effectively addressed by BO method, which employs a GP model to approximate the unknown function $J(\cdot)$. BO iteratively selects candidate points for evaluation and uses the gained information to guide the search towards the optimal solution. However, implementing this process in a physical robotics setup is typically troublesome, since each direct evaluation of $J(\cdot)$ on the robot can be resource-intensive, potentially causing wear and taking a long time to execute. 

To minimize the number of physical evaluation queries and optimize the parameters efficiently, we leverage a simulator model of the quadruped robot~\cite{ji2022synthesizing}. This simulator model serves as a digital replica of the real robot, providing evaluation results with reduced effort and time compared to physical evaluations. Consequently, the optimal combination of parameters can be identified through the optimization of the evaluation results obtained via simulation, making the process more cost-effective.

Nonetheless, it is crucial to account for the unavoidable discrepancies between the simulator and the real robot, 
and the non-negligible reality gap is demonstrated in our previous experiments~\cite{ji2022synthesizing}. As an alternative, the optimal points can be identified directly from the physical robot with the BO framework; however, as previously discussed, $J(\cdot)$ is expensive to evaluate in the physical experiments, resulting in increased workload and difficulty in data collection during the iterative process. 

To ensure that optimal performance can be achieved in the physical environment, we introduce an MFBO approach based on the BO framework for domain adaptation and online training. It facilitates the transfer of knowledge from simulation to reality during parameter optimization. Thus, the optimal solution is adapted and evolved efficiently from model-based optimization to the physical implementation.

\subsection{Multi-fidelity Bayesian Optimization}
\label{sec:MFBO}




Both the simulator and the physical robot can be modeled as two different GP models with the same input and output states. Thus, in the BO framework, the two GP models for the same subject are correlated, and their inter-model correlations can be exploited for efficient learning.
To achieve this, we employ the MTGP model to substitute the single output GP model. This substitution allows us to leverage multiple outputs that capture diverse behaviors of various plant models. We create a tuple $\langle \mathbf{p}_{i}, j\rangle$ by augmenting the input sample with the fidelity identifiers $j$, where $j \in \{1, 2\}$ denotes the simulator and physical environment. 
To reflect the similarity constraint, we define a new covariance function over the identifier by $k_f(j, j')$. With \emph{intrinsic model of coregionalization}~\cite{alvarez2012kernels}, we reconstruct the kernel function with
\begin{equation}
\label{eq:kernel_MTGP}
    \Tilde{k}(\langle\mathbf{p}_{i}, j\rangle, \langle\mathbf{p}_{i'}, j'\rangle) = k_f(j, j') k(\mathbf{p}_{i}, \mathbf{p}_{i'}),
\end{equation}
where $k(\cdot,\cdot)$ is defined in Sec.~\ref{sec:GP}. $k_f(\cdot,\cdot)$ is the additional covariance function representing the correlation over models, which takes the model labels as input.

For the training input set $\mathbf{P}$ with $n$ samples, we first assume the complete set of evaluation results on $J(\cdot)$ is available for both fidelity levels with $\Tilde{\mathbf{v}} = [\mathbf{v}^1, \mathbf{v}^2]$, where $\mathbf{v}^j = [v_1^j,\!\dots\!,v_n^j], j \in \{1, 2\}$.
Based on (\ref{eq:kernel_MTGP}), 
the covariance matrix $\Tilde{K}$ for the data from both fidelity levels can be obtained with
\begin{equation}
\label{eq:MT_covariance_matrix}
    \Tilde{K} = K_f \otimes K_\mathbf{P},
\end{equation}
where $\otimes$ denotes Kronecker product, $K_f \in \mathbb{R}^{2\times2}$ is a symmetric matrix whose diagonal entries represent the inner-task relationship with themselves, and the other elements describe the correlations of different fidelity levels. Note that if all the off-diagonal entries are zeros, $K_f$ becomes an identity matrix, which means all the underlying data sources are independent. 

 
The parameters for an MTGP model include $\pmb{\lambda}$ and $K_f$. We denote them collectively as $\Lambda\!\!=\!\!\begin{bmatrix}\pmb{\lambda}&\textrm{vec}(K_f)\end{bmatrix}^\top$, where $\textrm{vec}(\cdot)$ denotes the column vectorization of a matrix.
An MTGP model is optimized similarly to that in Sec.~\ref{sec:GP} by maximizing the marginal likelihood for $\Lambda$. We leave out the derivation of hyperparameter learning for simplicity. 
Accordingly, the prediction of the test point $\langle\mathbf{p}', j'\rangle$ can be achieved by calculating the conditional probability with
\begin{equation}
\label{eq:posterior_likelihood_MTGP}
    p(v' \mid \langle\mathbf{p}', j'\rangle, \Lambda^*, \mathbf{P}, \Tilde{\mathbf{v}}) = \mathcal{N}(\mu_{\mathbf{p}'}, \sigma_{\mathbf{p}'}),
\end{equation}
and
\begin{equation}
\label{eq:mean_covariance_expression_actuator_MTGP}
\begin{aligned}
    &\mu_{\mathbf{p}'} = \Tilde{k}(\mathbf{P}, \langle\mathbf{p}', j'\rangle)^\top \Tilde{K}^{-1} \Tilde{\mathbf{v}}^\top,\\
    &\sigma_{\mathbf{p}'} = \Tilde{k}(\langle\mathbf{p}', j'\rangle, \langle\mathbf{p}', j'\rangle)-\Tilde{k}(\mathbf{P}, \langle\mathbf{p}', j'\rangle)^\top \Tilde{K}^{-1} \Tilde{k}(\mathbf{P}, \langle\mathbf{p}', j'\rangle),   
\end{aligned}
\end{equation}
where $\Tilde{k}(\mathbf{P}, \langle\mathbf{p}', j'\rangle)$ denotes the kernel vector in the MTGP model for the test point tuple $\langle\mathbf{p}', j'\rangle$, and
\begin{equation}
    \label{eq:test_point_MTGP_kernel_matrix}
    \Tilde{k}(\mathbf{P}, \langle\mathbf{p}', j'\rangle) = (K_f)_{j'} \otimes k(\mathbf{P}, \mathbf{p}').
\end{equation}
$(K_f)_{j'}$ is a $2 \times 1$ vector where $(\cdot)_{j'}$ selects the $j'$-th column of the matrix $K_f$. Recall that $k(\mathbf{P}, \mathbf{p}')$ is an $n \times 1$ vector, which leads to the size of kernel vector $\Tilde{k}(\mathbf{P}, \langle\mathbf{p}', j'\rangle)$ becoming $2n \times 1$. It evaluates the kernel value of the test input with respect to all training data points from both fidelity models. To differentiate the single output GP model in (\ref{eq:posterior_likelihood})-(\ref{eq:mean_covariance_expression_actuator}), we denote by $\Tilde{g}(\langle\mathbf{p}, j\rangle)$ the MTGP model.

Built upon the MTGP model, the MFBO process can be updated accordingly. To be specific, in (\ref{eq:EI_acquisiton})-(\ref{eq:EI_acquisition_fun}), $g(\mathbf{p})$ is evaluated instead by $\Tilde{g}(\langle\mathbf{p}, j\rangle)$ through (\ref{eq:kernel_MTGP})-(\ref{eq:test_point_MTGP_kernel_matrix}). And the generated point by maximizing the acquisition function corresponds to an optimal candidate point for the physical robot. By performing the MFBO iterations, the precision of the MTGP is enhanced by incorporating additional data from the physical measurement. Consequently, this process facilitates the identification of the optimal point for the physical robot, thereby enabling the achievement of domain adaptation across models efficiently.

\begin{rem}
Owing to the expression simplicity, we utilize the \emph{isotopic data}~\cite{alvarez2012kernels} for the derivation in (\ref{eq:MT_covariance_matrix})-(\ref{eq:test_point_MTGP_kernel_matrix}), where the outputs $\Tilde{\mathbf{v}}$ are assumed fully observable for both simulator and physical robot based on the input set $\mathbf{P}$. However, in practical implementation, the collected data $\Tilde{\mathbf{v}}$ cannot be uniform among different fidelity levels due to the extraordinary cost of the physical experiment. It leads to the \emph{heterotopic data} scenario, where some of the outputs from the physical environment are unavailable. Nonetheless, the derivation still holds with the heterotopic data, but the covariance matrix $\Tilde{K}$ is inappropriate with matrix-wise calculation, as some elements in $K_\mathbf{P}$ of (\ref{eq:MT_covariance_matrix}) correspond to un-examined data and will not be observed. Alternatively, if the number of data is $n^j$ for the $j$-th data source, the matrix size will be changed, and $\Tilde{K}$ becomes an $N\times N$ matrix, where $N = n^1 + n^2$. Each entry in $\Tilde{K}$ can still be obtained by using (\ref{eq:kernel_MTGP}) through element-wise calculation. Eqs. (\ref{eq:posterior_likelihood_MTGP})-(\ref{eq:test_point_MTGP_kernel_matrix}) will change accordingly, where $\Tilde{k}(\mathbf{P}, \langle\mathbf{p}_{t}, j\rangle)$ becomes a vector of size $N \times 1$.
\end{rem}

The MFBO approach can be outlined as a two-stage learning process, as summarized in \textbf{Algorithm 1}. First, we initialize a training data set by randomizing a set of parameter candidates $\mathbf{P}_\textrm{init}$ and measuring their output $\mathbf{v}_\textrm{init}$ through simulation. They construct the data set for the initial GP model training. Subsequently, by running the traditional BO on the simulation platform (with the fidelity identifier $j\!=\!1$) in an iterative way, we generate an optimal parameter vector $\mathbf{p}_{j=1}^*$, along with the accumulation of data $[\langle\mathbf{p}_i, j\!=\!1\rangle, v_i]$. After that, we leverage $\mathbf{p}_{j=1}^*$ as the starting point and perform MFBO on the physical experiment platform (with the fidelity identifier $j\!=\!2$). Following the evaluation of each candidate point $\mathbf{p}_{k}$, we augment the data set with a new tuple $\langle\mathbf{p}_{k}, j=2\rangle$ and its corresponding result $v_{k}$. The optimal parameter $\mathbf{p}_{j=2}^*$ is determined upon the completion of iterations in the MFBO process.

Importantly, by using the MFBO approach in Lines 12-23, the amount of information gained from both sources with different fidelities can be used and contribute to the optimization process on the real robot, which saves a significant overhead by avoiding re-exploring the same parameter space on the physical platform. Owing to the computational efficiency and accuracy of the simulation platform, our MFBO approach yields substantial improvements in terms of the speed of optimization.

\begin{algorithm}[tbh]
\KwData{Initial set $\mathbf{P}_\textrm{init}$, design space bounds $[\underline{\mathbf{p}_i},\overline{\mathbf{p}_i}]$, initial guess of $K_f$, number of iterations $i_{\mathrm{max}}$, $k_{\mathrm{max}}$, $i=k=1$}
\KwResult{Optimal parameter on real robot $\mathbf{p}_{j=2}^*$}
Evaluate $\mathbf{P}_\textrm{init}$ on the simulator and get $\mathbf{v}_\textrm{init}$\;
Build $\mathcal{T} \leftarrow \{[\langle\mathbf{p}_i, j=1\rangle, v_i] \mid \mathbf{p}_i \in \mathbf{P}_\textrm{init}, v_i \in \mathbf{v}_\textrm{init}\}$\;
Train GP model $g(\cdot)$ from $\mathcal{T}$ with parameter $\pmb{\lambda}$\;
\While{$i<i_\mathrm{max}$}{
Perform BO based on $\mathcal{T}$ and $\pmb{\lambda}$, get new candidate point $\mathbf{p}_i$\;
Evaluate $\mathbf{p}_i$ on the simulator and get $v_i$\;
Update set: $\mathcal{T} \leftarrow \mathcal{T} \bigcup \{[\langle\mathbf{p}_i, j=1\rangle, v_i]\}$\;
Update the GP model $g(\cdot)$ and get a new $\pmb{\lambda}$\;
Update iteration: $i \leftarrow i+1$;}

$\mathbf{p}_{j=1}^* \leftarrow \mathbf{p}_i, i = \arg\max_i(v_i), [\langle\mathbf{p}_i, j=1\rangle, v_i] \in \mathcal{T}$\;

\While{$k<k_\mathrm{max}$}{
\uIf{$k=1$}{$\mathbf{p}_k=\mathbf{p}_{j=1}^*$, $\Lambda=\begin{bmatrix}\pmb{\lambda}&\textrm{vec}(K_f)\end{bmatrix}^\top$}
\Else{Perform MFBO based on $\mathcal{T}$ and $\Lambda$, get a new candidate point $\mathbf{p}_k$\;}
Evaluate $J(\mathbf{p}_k)$ on the real robot and get $v_k$\;
Update set: $\mathcal{T} \leftarrow \mathcal{T} \bigcup \{[\langle\mathbf{p}_k, j=2\rangle, v_k]\}$\;
Update the MTGP model $\Tilde{g}(\cdot)$ and get a new $\Lambda$\;
Update iteration: $k \leftarrow k+1$;}
$\mathbf{p}^* \leftarrow \mathbf{p}_k, k = \arg\max_k(v_k)$, $[\langle\mathbf{p}_k, j=2\rangle, v_k] \in \mathcal{T}$\;

\caption{Multi-fidelity Bayesian optimization}\label{algorithm_MFBO}
\end{algorithm}

\section{System Architecture Overview}
\label{sec:experiment_setup}
To run the MFBO algorithm on the physical robot, the algorithm requires either real-time access to the robot simulation model or the storage of the training data for solving the BO problem of (\ref{eq:BO_objective_fun}). Either is a big challenge to the training efficiency and the real-time control of the physical robot. To alleviate the data storage and computational burden on the robot's onboard microprocessor, we employ an edge computing framework to offload the MFBO training process to an edge server.

\begin{figure}[!t]
	\centering
		\includegraphics[width=3in]{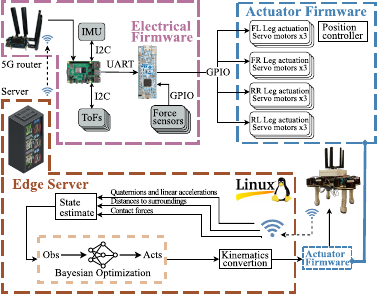}
	\caption{Control system architecture. The actuator and electrical firmware are physically integrated into the robot. The unidirectional arrows surrounding radio waves denote the transmission and reception of messages between soft robots and the edge server.} \label{fig:architecture}
\end{figure}

The control system architecture of the soft quadruped robot is illustrated in Figure~\ref{fig:architecture}. 
The robot's onboard computation task mainly focuses on the sensor data collection and control signal actuation. This corresponds to a low level of decentralization (sometimes referred to as a distributed I/O approach). Three types of sensors are integrated into the robot, including the Inertial Measurement Unit (IMU), Time of Flight (ToF), and force sensors. Given the study's emphasis on walking speed, both IMU and ToF sensors are employed to estimate longitudinal velocity, while force sensors only contribute to footfall pattern analysis. The onboard collected data from the sensors are sent to the edge server for the state estimation of the robot's velocity. Subsequently, the robot receives control signals from the edge server to govern soft leg actuation. Actuation is executed through position control of the servo motors, with lower-level PD controllers enhancing stability and efficiency.

The edge server is dedicated to executing computationally and memory-intensive tasks, encompassing functions such as state estimation, MFBO training, CPG oscillator calculation, and kinematics conversion. All functionalities on the edge server are deployed within a virtual machine operating in the Ubuntu 20.04 environment. The raw data of sensor feedback from the soft quadruped robot, including quaternions and linear accelerations, distances to surroundings, and contact forces, are fused to enhance speed estimation through a Kalman filter. The resultant stabilized speed signal serves as the input for MFBO training, facilitating the optimization of parameters for the CPG controller. Subsequently, the oscillator signal is updated, and actuation commands are derived through kinematics conversion. The communication infrastructure between the robot's controller and the edge server relies on 5G technology, facilitated by the Quectel RM500Q-GL 5G HAT integrated into the soft robot. The employment of 5G communication not only provides improved reliability and lower latency for the real-time control task, but also enhances the mobility support with expanded coverage, which makes it particularly advantageous for potential industrial IoT use cases with the soft robots.  

\section{Results and Validation}
\label{sec:result_validation}
\subsection{Baseline Selection}
To assess the effectiveness of the implemented BO approach during the training phase, we conduct performance comparisons with the following baselines. These baselines include algorithms/search methods previously introduced in Sec.~\ref{sec:introduction}, which serve as a reference for state-of-the-art approaches in the domain.
\begin{itemize}
    \item Adaptive grid search (AGS)~\cite{shi2022adaptive}: Owing to the exponential complexity associated with grid search at a finer granularity for sampling, an Adaptive Grid Search (AGS) approach is used to avoid exhaustive search. AGS iteratively refines the search space based on promising regions from the previous optimal solution. We consider the point identified by AGS as the global optimum.
    \item Simulated annealing (SA)~\cite{delahaye2019simulated}: SA is utilized as the benchmark for black-box optimization comparison due to its effectiveness as a single-point search algorithm. Note that population-based algorithms, e.g., Genetic Algorithms (GA)~\cite{wang2020transfer} and Particle Swarm Optimization (PSO), are not ideal due to the computational expense of evaluating a substantial amount of samples per iteration.
    \item Random search (RS)~\cite{andradottir2006overview}: RS represents a worst-case scenario without intelligent exploration, which is used to quantify the improvement of the other methods.
    \item Omni-directional walking design (OWD)~\cite{ji2022omnidirectional}: OWD is proposed in the previous work for the soft robot walking gait design. It is regarded as a baseline in terms of ad hoc design without optimization or search processes; thus, it is only used to compare the performance of the identified optimal gait pattern.
\end{itemize}

 Note that, except OWD, the other baselines are solely employed for evaluating the optimization performance at the simulation level, 
 as they cannot effectively control the physical robot to walk. 
 Except AGS, each search method is assessed through 3 evaluations using distinct random seeds. We impose a predefined limit of 300 iterations as a stopping criterion for SA, RS, and BO. 

 The selection of the surrogate model in BO is crucial in identifying optimal points of design parameters. 
 As commented in Sec.~\ref{sec:Bayesian_opt}, various surrogate models may be employed for BO. Specifically, we assess the performance of these surrogate models against GP, which are summarized and outlined in a previous study~\cite{bliek2023benchmarking}:

 \begin{itemize}
 
     \item Random forest (RF)~\cite{hutter2011sequential}: RF is an ensemble method consisting of multiple regression trees. Each tree is trained on a bootstrap sample of the training data, and the final prediction is obtained by averaging the predictions of all trees.
     \item Random Fourier Expansion (RFE)~\cite{bliek2016online}: RFE generates random features using Fourier basis functions and trains a linear regression model on these features to approximate the objective function.
     \item Support Vector Machine (SVM)~\cite{saves2024smt}: SVM finds the best hyperplane to separate different classes by maximizing the margin between support vectors. We use Radial Basis Function (RBF) as the kernel function.
     \item Piece-wise Linear (PL)~\cite{bliek2021black}: PL surrogate model uses rectified linear units (ReLUs) in a linear combination, where the ReLU input represents the linear mapping of the input data~\cite{bliek2021black}.
 \end{itemize}
 
 The hyperparameters in the surrogate models are configured in alignment with the studies referenced by Bliek et al.~\cite{bliek2023benchmarking}. All the simulation evaluations are executed on a desktop with Intel(R) Core(TM) i7-8700 CPU at 3.20 GHz, six cores, and 8.0 GB installed memory (RAM).

\subsection{Comparison of Optimization Methods}
Table~\ref{table:comp_benchmark} lists the optimization results by different methods with the associated computational time. Note that this comparison is conducted with the simulation platform. Among all the methods, ADS delivers the best performance with the global optimum speed at 0.312 $m/s$. Nevertheless, it comes at a significant computational time cost, consuming over 72 hours as it evaluated 3072 points within 3 iteratively refined parameter regions. In contrast, the BO approach only takes 1.278 hours to converge after roughly 70 iterations, reaching an optimal speed of 0.286 $m/s$, which corresponds to 91.67\% of the global optimum. In addition, SA does not meet the objective function tolerance criteria and gets terminated at 300 rounds. It results in an identified optimal point of 0.264 $m/s$ in a duration of 5.025 hours, which is slightly worse than BO. On the other hand, RS provides a lower bound reference of employing the defined CPG oscillator-based gait pattern. It offers a baseline speed of 0.125 $m/s$ and consumes 4.567 hours to finish the 300-iteration search. In comparison to ADS and RS, the application of the BO approach demonstrates its effectiveness in exploring the optimal gait pattern for soft robots. Additionally, in the context of black-box optimization, it also exhibits superior efficiency compared to SA. Moreover, the underlying GP model in BO facilitates domain adaptation for subsequent sim-to-real transfer, which is a feature not achievable by SA.
\begin{table}[!t]
\centering
\caption{Comparison of the optimization performance}
\label{table:comp_benchmark}
\begin{adjustbox}{width=\linewidth}
\begin{tabular}{r|ccccc} 
\hline
\multicolumn{1}{l}{} & ADS & SA & RS & BO \\ 
\hline
Max speed ($m/s$)& 0.312& 0.264& 0.125& 0.286 \\
Optimization time ($h$)& 72.367& 5.025 & 4.567 & 1.278 \\
\hline 
\end{tabular}    
\end{adjustbox}
\end{table}

\begin{figure}[!t]
	\centering
		\includegraphics[width=3in]{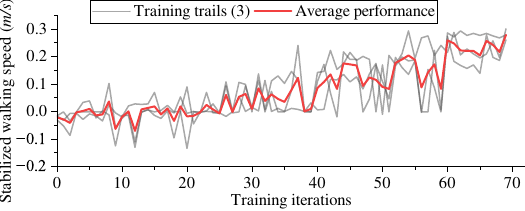}
	\caption{BO training on simulation platform} \label{fig:BO_training}
\end{figure}

For a more comprehensive analysis of the robustness of BO, Figure~\ref{fig:BO_training} illustrates the BO training performance across different trials with randomly initialized seeds and noise, where the corresponding parameter setting is given in Table~\ref{table:param_setting}. Note that the variable ranges are defined with the same extreme values as deployed in the study~\cite{ji2022omnidirectional} for a fair comparison. It can be seen that the robot stays at an exploration phase within the design space for the first 40 iterations, followed by a noticeable performance improvement for the next 20 iterations. Subsequently, convergence is achieved with a stable speed at around 70 iterations. In the simulated environment, the robot ultimately learns a stable speed of approximately 0.28 $m/s$. 
This represents a remarkable improvement compared with the previous study~\cite{ji2022omnidirectional} (0.14 $m/s$). The optimized gait pattern $\mathbf{s}$ within the identical motion space leads to a twofold increase in walking speed.

\subsection{Comparison of Surrogate Models}
\label{sec:surrogate_comp}
Table~\ref{table:comp_surrogate} compares the performance of different surrogate models. We construct a test data set by pseudo-randomly selecting 1000 samples in order to cover the entire design search space.
We evaluate goodness-of-fit of models with $R^2$ and NMSE to quantify the prediction accuracy. To evaluate the generalization and robustness of the model, we conduct the experiments through a 5-fold \emph{cross-validation} by partitioning the dataset into five subsets, using each subset once as a validation set, and averaging the performance metrics across all folds. Hence, we present the mean $R^2$ and its standard deviation from cross-validation, along with the mean NMSE. In addition, we evaluate the computational time for each surrogate model regression to assess their computational complexity.

In Table~\ref{table:comp_surrogate}, it is evident that under a relatively small dataset scenario, the employed GP surrogate model exhibits a noticeable advantage over other methods in terms of modeling accuracy. Furthermore, during the cross-validation experiment, it demonstrates the lowest standard deviation value, indicating stable out-of-sample performance and strong generalization to unseen data. On the other hand, the GP method has larger computational complexities, which grow cubically and quadratically with the number of training points during model training and online inference, respectively. In comparison to other benchmark methods, it only shows better computational efficiency against PL, which requires back-propagating training for ReLUs. This result gives us insights into the trade-off between model accuracy and real-time efficiency when employing the GP model, thereby emphasizing the significance and justification of our edge off-loading architecture.

\begin{table}[!t]
\centering
\caption{Cross-validation for comparing the surrogate models}
\label{table:comp_surrogate}
\begin{adjustbox}{width=\linewidth}
\begin{tabular}{r|ccccc} 
\hline
\multicolumn{1}{l}{} & RF & RFE & SVM & PL & GP \\ 
\hline
Mean $R^2$         & 0.9373 & 0.8880 & 0.8902 & 0.9294 & 0.9856\\
Standard deviation & 0.0121 & 0.0262 & 0.0220 & 0.0140 & 0.0049\\
Mean NMSE          & 0.1134 & 0.1519 & 0.1487 & 0.1201 & 0.0478\\
Computational time ($s$)    & 0.0546 & 0.1584 & 0.1180 & 10.9937 & 4.7888\\
\hline 
\end{tabular}    
\end{adjustbox}
\end{table}


\begin{table}
\centering
\caption{Parameter setting for BO training}
\label{table:param_setting}
\begin{tabular}{cc} 
\hline
\multicolumn{1}{c}{Parameters}     & Values                             \\ 
\hline
$\mid\mathbf{P}_\textrm{init}\mid$ & 10                                 \\
$\overline{\mathbf{p}}$            & $[0.7, 0.008, 2, 0.9, 2\pi]^\top$  \\
$\underline{\mathbf{p}}$           & $[0.1, 0.001, 0.4, 0.1, 0]^\top$   \\
$i_{max}$                            & 70                                 \\
\hline
\end{tabular}
\end{table}

\subsection{MFBO and Physical Deployment}
\begin{figure}[!t]
	\centering
		\includegraphics[width=3in]{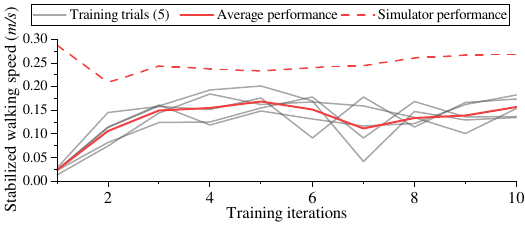}
	\caption{MFBO physical training} \label{fig:physical_training_trails}
\end{figure}

This section addresses the results of MFBO during the physical deployment, and the training process is depicted in Figure~\ref{fig:physical_training_trails}. As previously noted, we initiated the MFBO training with the optimal gait discovered in the simulation environment. Due to model errors, the robot's actual walking behavior differs from the simulated speed. Specifically, it struggles to attain the desired speed and only reaches approximately 0.025$m/s$\footnote{\label{footnote1}The corresponding motion patterns are viewable in the attached video.}. Nevertheless, after several iterations of MFBO training, the robot rapidly adjusts the walking pattern to suit the physical environment. Consequently, its walking speed gradually increases to around 0.2 $m/s$ and reaches a saturation point. Despite some oscillations and fluctuations in the training curves across 5 different trials attributable to environmental noise, they exhibit a consistent overall trend and stabilize within 10 iterations.

\begin{figure}[!t]
	\centering
		\includegraphics[width=3in]{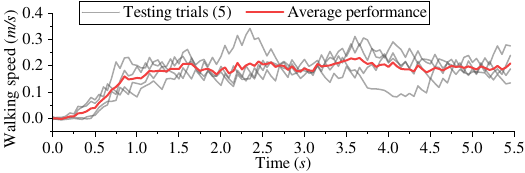}
	\caption{Walking test of the optimized parameters.} \label{fig:physical_test_max_spd}
\end{figure}

\begin{figure}[!t]
	\centering
		\includegraphics[width=3in]{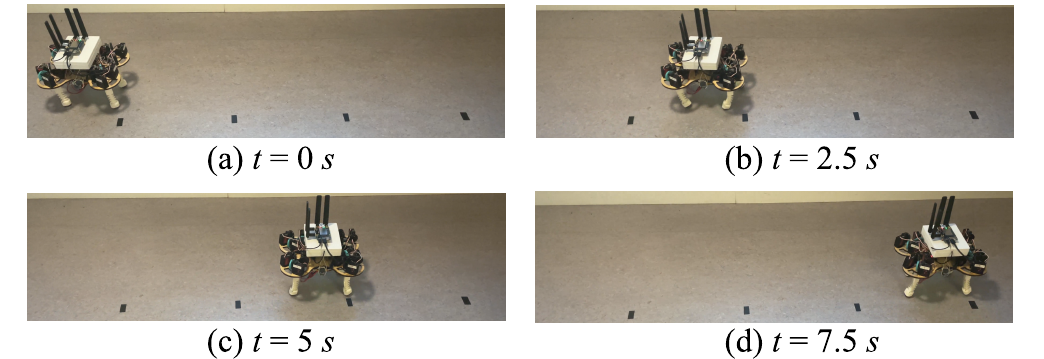}
	\caption{Physical test results of the converged motion pattern captured in video frames.} \label{fig:video_figs}
\end{figure}

Furthermore, the robot's walking speed at the optimized point in the physical environment is plotted in Figure~\ref{fig:physical_test_max_spd}. As before, we conduct 5 trials and provide both individual trial results and the average performance. The robot accelerates to a stable speed with approximately 1.2 $s$. In particular, the largest speed goes up to 0.344 $m/s$ during the testing with an average stabilized speed of 0.214 $m/s$. This represents a significant improvement of 52.71\% compared to the physical testing results of the OWD solution within the same parameter range. Figure~\ref{fig:video_figs} shows the physical test results of the converged motion pattern\footref{footnote1}.

\begin{figure}[!t]
	\centering
		\includegraphics[width=3in]{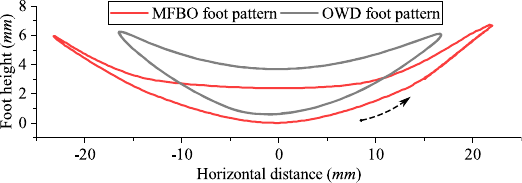}
	\caption{Comparison of the foot pattern.} \label{fig:xz_displacement}
\end{figure}

Figure~\ref{fig:xz_displacement} plots and compares the foot trajectories between the MFBO and OWD gaits. The MFBO controller demonstrates a substantially increased swing distance in the direction of motion, resulting in an approximately twofold expansion of the horizontal distance compared to the OWD foot pattern. In addition, both foot patterns attain a similar maximal foot height. Note that the design space of the leg's state $\mathbf{s}$ remains consistent across both methods. However, the swing trajectories of $\alpha_b$ and $z_l$ are different for two methods during the design phase. This disparity leads to distinct phase synchronization behaviors between $\alpha_b$ and $z_l$, influencing the maximal horizontal distances and subsequently impacting walking patterns and speeds.

\begin{figure}[!t]
	\centering
		\includegraphics[width=3in]{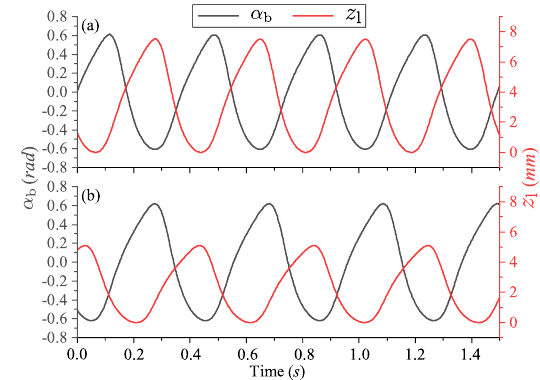}
	\caption{Comparison of behaviors of optimized oscillators. (a): Simulation platform, (b): Physical testing platform.} \label{fig:oscillator_comp}
\end{figure}

In addition, as shown in Figure~\ref{fig:oscillator_comp}, a comparison of oscillator behaviors is made between the optimal point identified in the simulator and the physical environment. According to (\ref{eq:oscillator_to_cycle}), the oscillator exhibits varying characteristics depending on the design parameter vector $\mathbf{p}$. Relative to the optimized gait observed in the simulator, the oscillator associated with the optimal gait in the physical test exhibits a comparatively lower frequency value $f$ and a higher value for the shape ratio $\alpha$. These differences suggest a more prompt transition between the two swing phases during physical training. The maximal bending angle $\alpha_\textrm{b}$ remains consistent across both data sources; however, in the physical testing scenario, the robot's leg requires less time for the swing-back motion and tends to exert a greater force during the rearward push against the ground. 
This phenomenon corresponds to the previously discussed reality gap in Figure~\ref{fig:physical_training_trails}, as the robot encounters challenges in generating sufficient power to facilitate forward movement when operating with the optimal gait derived from the simulator. Consequently, it shows more aggressive behavior by shortening the duration of the swing-back phase. Moreover, in Figure~\ref{fig:oscillator_comp}(b), $z_\textrm{l}$ exhibits a noticeable decrease accordingly, which is inferred to compensate for the alteration in $\alpha$, since keeping the same amount of $z_\textrm{l}$ with a larger $\alpha$ will challenge the motor response capacity. Also, increasing $\alpha$ without decreasing $z_\textrm{l}$ will result in an overly aggressive gait pattern, potentially leading to kinetic friction and causing slippage.

\subsection{Off-loading performance comparison}
To highlight the necessity of edge computing architecture to realize the task off-loading process, we conducted a comparative analysis of CPG calculations executed on the robot's control board and the edge server. Note that the MFBO training part is exclusively offloaded to the edge server and is not considered in this comparison due to its excessive computational resource demands on the robot's board. Owing to the contention of computational resource in a multi-thread setup, control tasks on the robot's onboard controller may become overdue, resulting in missed deadlines. Therefore, we evaluate the overdue status of the control tasks.

\begin{figure}[!t]
	\centering
		\includegraphics[width=3in]{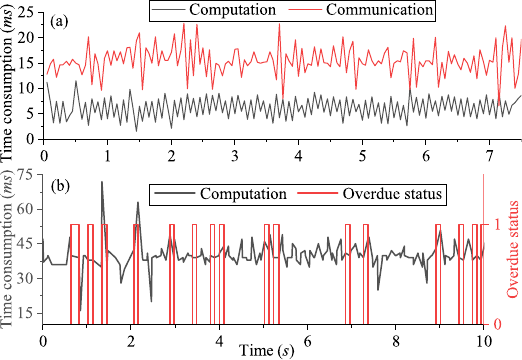}
	\caption{Comparison of the time consumption of real-time control when the CPG is calculated at (a): edge server, (b): robot's onboard controller.} \label{fig:offload_onboard}
\end{figure}

As shown in Figure~\ref{fig:offload_onboard}(a), when the control task is offloaded, the computation of the CPG actuation signals is accelerated by leveraging the advanced computation capacity of the edge server. In this scenario, 5G communication contributes to a larger portion of the total delay, which is primarily limited by the fixed proximity of the edge server and 5G bandwidth availability. The resulting end-to-end latency satisfies the defined control frequency (20 Hz), thus enabling a smooth gait actuation in the physical deployment, with average and maximum delays 28.215 $ms$ and 42.779 $ms$, respectively. Conversely, executing the computation task on the onboard controller leads to longer execution time delays, and some jobs miss the deadline. We profile the overload of the onboard computation from a thread in the lower-level controller in Figure~\ref{fig:offload_onboard}(b) and evaluate the performance influence. It is noteworthy that the plotted time consumption in Figure~\ref{fig:offload_onboard}(b) exclusively reflects the execution of the control task. However, some instances of missed deadlines, as observed in the figure, occur despite computation delays below 50 $ms$. This phenomenon is attributed to the incomplete execution of perception and actuation tasks within the closed-loop control period. Thus, the degradation in real-time performance results in an inconsistent and incoherent gait pattern, with the robot getting stuck in certain postures due to missed actuation signals\footnote{The comparison of real-time control performance is shown in the video attachment.}.

\subsection{Domain Adaptation and Reality Gap}
As discussed in Sec.~\ref{sec:MFBO}, the domain adaptation of MFBO is realized by determining the matrix $K_f$ with physical experiments, and it represents the correlation between the simulator and the physical robot. To increase its interpretability, a transformation is performed on $K_f$ by \emph{diagonal elements normalization}~\cite{durichen2014multitask}. We refer to the normalized matrix as $K'_f$, and
    $K'_f = \begin{bmatrix}
        1 & 0.575\\
        0.575 & 1
    \end{bmatrix}$.
The diagonal elements of $K'_f$ are unit values to eliminate the influence of label-specific scaling. The off-diagonal elements indicate that the dynamic properties of the simulator and the robot have a moderately positive correlation.

In addition, we perform a bi-directional comparison to verify the reality gap, where the performance of the optimized parameters in different MFBO iterations from a training trail is cross-validated in the simulator. The corresponding velocities are plotted in Figure~\ref{fig:physical_training_trails}. Notably, in the simulated platform's initial iteration, the robot achieves its maximum velocity despite exhibiting the worst physical performance. 
Subsequently, the velocities in the simulator drops and gets stabilized for the other candidate points as the MFBO physical training progresses. Nevertheless, it consistently maintains a higher velocity than that observed in the physical realization. The cross-validation verifies the non-linear correlation of the addressed objective function across different data sources. Based on the simulation results, the \emph{Pearson's correlation coefficient} is calculated with 
$K_\textrm{bi} = \begin{bmatrix}
        1 & 0.601\\
        0.601 & 1
\end{bmatrix}$. The high consistency between $K_\textrm{bi}$ and $K_f$ verifies that our estimated model has a high accuracy for domain adaptation, which leads to the realization of the MFBO with high efficiency.

\section{Conclusion and Future Work}
\label{sec:conclusion}
This study investigates advanced gait pattern design for a tendon-driven soft quadruped robot, with a specific focus on bridging the reality gap mitigation through learning-enhanced methods. The gait design exploits a parametric model based on Central Pattern Generators (CPG) with Hopf oscillators, whose parameters are determined through Bayesian Optimization (BO) using both simulation and experimental data. Furthermore, to ensure a smooth sim-to-real transfer, we refine the BO method and propose a multi-fidelity Bayesian Optimization (MFBO) approach for domain adaptation, thus efficiently compensating for modeling errors and overcoming the reality gap. Moreover, the entire training process utilizes an edge computing off-loading scheme with 5G communication, enhancing real-time data collection, training, control performance, and potential industrial applications. Future work includes investigation into walking properties for optimization and exploration of high-level navigation tasks on the soft quadruped robot platform. In addition, the communication dependence introduced by the off-loading scheme raises potential vulnerability due to communication loss. Hence, future work could extend this study by developing methods to enable the system to continue operation in a degraded mode and transition into a fail-safe state.

\section*{Declaration of competing interest}
The authors declare that they have no known competing financial interests or personal relationships that could have appeared to influence the work reported in this paper.

\section*{Declaration of generative AI and AI-assisted technologies in the writing process}
During the preparation of this work the author(s) used ChatGPT in order to improve readability and language of the work. After using this tool/service, the author(s) reviewed and edited the content as needed and take(s) full responsibility for the content of the publication.

\section*{Acknowledgements}
This research has been carried out as part of the TECoSA Vinnova Competence Center for Trustworthy Edge Computing Systems and Applications at KTH Royal Institute of Technology (www.tecosa.center.kth.se) and is also part funded by the Entice project (Vinnova Advanced Digitalization program). Lei Feng is also partly funded by KTH XPRES. We acknowledge Gianfilippo Fornaro for his contribution to setting up the edge computing communication system.

\section*{CRediT authorship contribution statement}

$\textbf{Kaige~Tan:}$ Conceptualization, Methodology, Software, Validation, Formal analysis, Investigation, Writing Original Draft, Visualization.
$\textbf{Xuezhi~Niu, Qinglei~Ji:}$ Methodology, Software, Investigation.
$\textbf{Lei~Feng, Martin~Törngren:}$ Supervision, Project administration, Funding acquisition, Writing-Review $\&$ Editing.

\bibliographystyle{elsarticle-num}
\bibliography{main}

\end{document}